# FAST AND ROBUST REGISTRATION OF AERIAL IMAGES AND LIDAR DATA BASED ON STRUCTURAL FEATURES AND 3D PHASE CORRELATION

Bai Zhu, Yuanxin Ye*, Chao Yang, Liang Zhou, Huiyu Liu, Yungang Cao

Faculty of Geosciences and Environmental Engineering, Southwest Jiaotong University, Chengdu 610031, China

**Commission II, WG II/2**

**ABSTRACT:**

Co-Registration of aerial imagery and Light Detection and Ranging (LiDAR) data is quilt challenging because the different imaging mechanism causes significant geometric and radiometric distortions between such data. To tackle the problem, this paper proposes an automatic registration method based on structural features and three-dimension (3D) phase correlation. In the proposed method, the LiDAR point cloud data is first transformed into the intensity map, which is used as the reference image. Then, we employ the Fast operator to extract uniformly distributed interest points in the aerial image by a partition strategy and perform a local geometric correction by using the collinearity equation to eliminate scale and rotation difference between images. Subsequently, a robust structural feature descriptor is build based on dense gradient features, and the 3D phase correlation is used to detect control points (CPs) between aerial images and LiDAR data in the frequency domain, where the image matching is accelerated by the 3D Fast Fourier Transform (FFT). Finally, the obtained CPs are employed to correct the exterior orientation elements, which is used to achieve co-registration of aerial images and LiDAR data. Experiments with two datasets of aerial images and LiDAR data show that the proposed method is much faster and more robust than state of the art methods.

## 1. INTRODUCTION

At present, the airborne Light Detection and Ranging (LiDAR) and aerial photogrammetry systems are the main sources for obtaining a large amount of earth observation data. Combining the 3D information contained in LiDAR data with the rich semantic information in aerial imagery plays an important role in many applications such as building extraction (Awrangjeb et al., 2013), Change detection (Qin and Gruen, 2014), 3D reconstruction (Wu et al., 2018), etc. Image registration aims to align two or more images, which is a prerequisite step to combining LiDAR data and aerial images. It is well known that LiDAR data are the discrete three-dimension(3D) point clouds, while aerial images are the continuous two-dimension(2D) optical images. These two types of data have significant differences in geometry and radiation, which make it quite challenging to achieve the precise registration between them. In last decades, researchers of remote sensing community develop a lot of image registration methods for aerial images and LiDAR data, which can be roughly divided into the following three types of methods (Peng et al., 2019).

The first category belongs to the 3D-3D registration method. The aerial images with multiple overlaps are used to generate 3D point clouds by detecting dense correspondences between them (Glira et al., 2019). The registration between the two-point clouds is used to replace the registration of the original LiDAR data and the aerial images (Harrison et al., 2008). However, such registration methods rely on correspondence detection of aerial image sequences and cannot handle the registration of one single aerial image and LIDAR data, and the process of generating dense 3D point clouds might yield errors. Moreover, the registration accuracy may degrade because there are few true correspondences between the two point clouds. The second one is the 3D-2D registration method, which is to establish a transformation function by selecting feature primitives between the two types of heterogeneous data to achieve registration (Zhang et al., 2015). Due to significant differences between LiDAR point cloud data and aerial images, it is quite difficult to extract common features between such data. In practice, most 3D-2D registration methods usually require to detect reliable features by manual (Rönnholm and Haggrén, 2012). Therefore, this type of method cannot effectively be applied for automatic registration of LiDAR data and aerial images.

The third category is the 2D-2D registration method, which interpolates a 3D LiDAR point clouds into a DSM, an intensity image, or a distance image, which can transform the 3D-2D registration into the 2D-2D registration. These methods can make use of existing algorithms from digital image registration (Zitova and Flusser, 2003), which are currently more mature and automatic than 3D-2D registration methods. These methods can be generally divided into intensity-based methods (Parmehr et al., 2012), feature-based methods (Lowe, 2004), and a combination of the previous two methods (Ye and Shan, 2014). However, most of these methods cannot effectively handle significant geometric distortions and intensity differences between aerial and LiDAR intensity images.

In order to address the issues, this paper proposes a registration method of aerial images and LiDAR data based on structural features and 3D phase correlation, which belongs to the 2D-2D registration method. Firstly, global geometric distortions such as scale and rotation differences between aerial image and LiDAR intensity images are eliminated by performing a local geometric correction, which is based on the collinearity equation with the position and orientation system (POS) data of aerial images. Secondly, structural features are extracted by using a robust descriptor named the Channel Feature of Oriented Gradient (CFOG) (Ye et al., 2019), and the 3D phase correlation is used as the similarity metric for detecting control points (CPs) by a template matching scheme. Subsequently, an outlier detection is carried out to remove mismatches based on the collinearity equation. Finally, we optimize the exterior orientation elements, and complete the registration of aerial images and LiDAR data.

* Corresponding author: Yuanxin Ye, yeyuanxin@swjtu.edu.cn

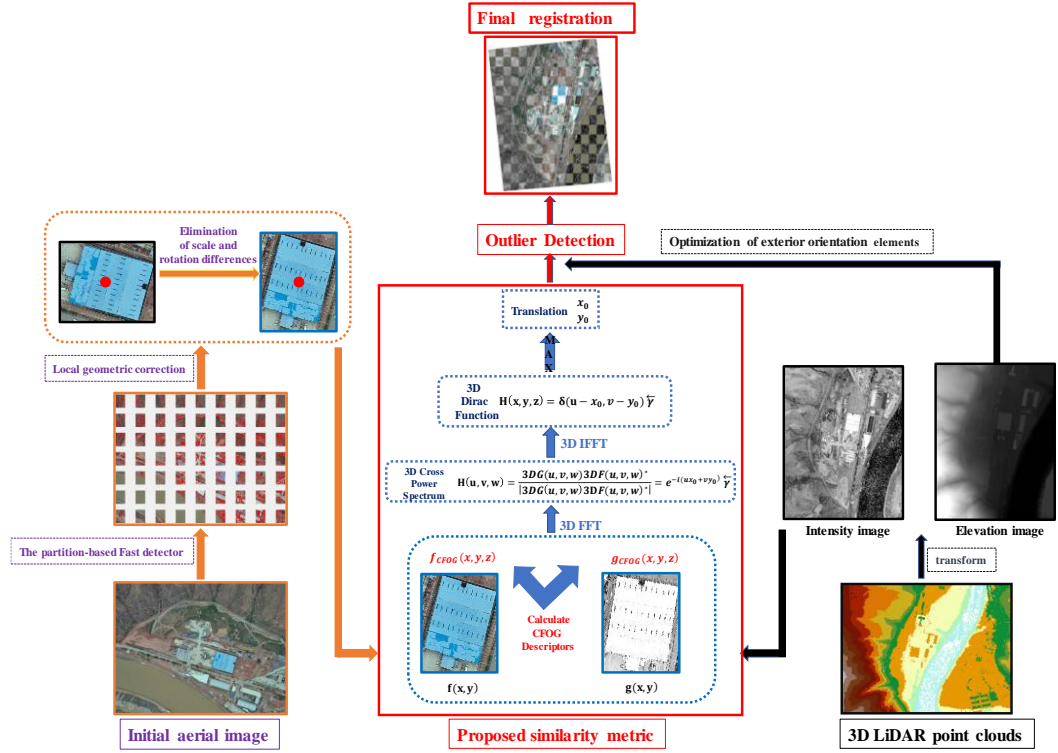

Figure 1. Flowchart of the proposed method

## 2. METHODOLOGY

This paper proposes an automatic registration method for aerial images with POS data and LiDAR intensity images. Figure 1 shows the flowchart of the proposed method, which is include four steps: (1) interest point extraction by a partition-based FAST detector; (2) local geometric correction based on the collinearity equation; (3) Image matching using the similarity metric based on structural features and 3D phase correlation; (4) exterior orientation elements optimization and outlier detection.

### 2.1 Interest point extraction

The extraction of interest points is the first step of the proposed method. Considering the reliability and computational efficiency of feature extraction, we use the Features from Accelerated Segment Test (FAST) detector (Rosten and Drummond, 2006) to detect interest points. However, it makes interest points unevenly distributed over the image (see Figure 2) when the FAST detector is directly applied for feature extraction, which is not beneficial to the following image registration.

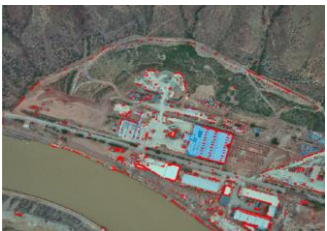

Figure 2. Interest point extraction by original FAST

To address that, we design a partition scheme for the extraction of interest points. Firstly, the image is divided into $n \times n$ non-overlapping grids, and we calculate the FAST value of each pixel for every gird. Then, the FAST value is sorted in order from large to small, and the k pixels with the largest FAST values are selected as the interest points. Figure 3 shows the interest points detected by the partition based FAST operator. Compared with the original one, the partition based FAST detector can extract more evenly distributed interest points in the image.

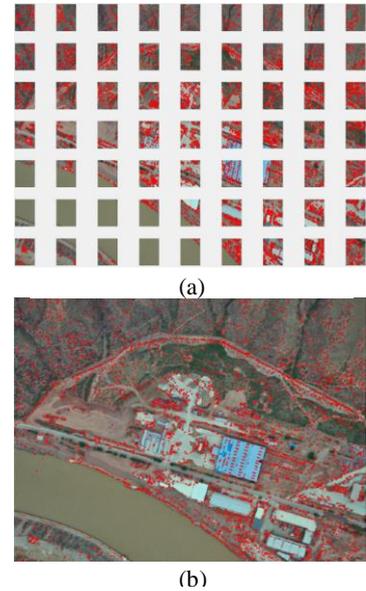

Figure 3. Interest point extraction by the partition-based FAST detector. (a)Non-overlapping grids based on the partition scheme. (b) Evenly distributed interest points in the image.

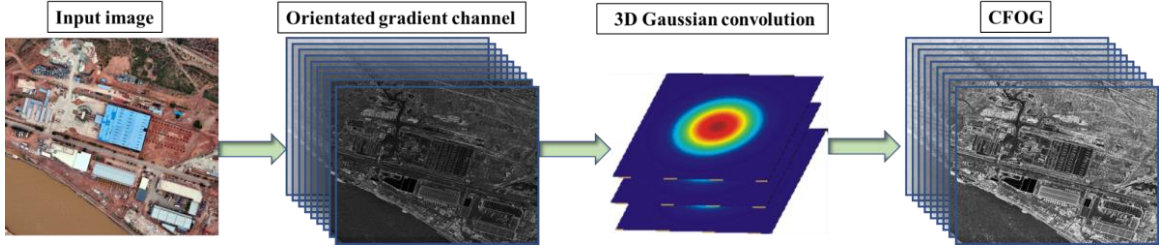

Figure 5. Construction of the CFOG Descriptor

**2.2 Local geometric correction**

Considering that the POS data of aerial images provide the initial exterior orientation elements, which can be used to coarsely correct the images. This process can effectively eliminate the rotation and scale differences between aerial images and LiDAR intensity images. Accordingly, we perform a local geometric correction on aerial images using the collinearity equation, which is expressed as:

$$\begin{cases} x = -f\frac{a_1(X-X_S)+b_1(Y-Y_S)+c_1(Z-Z_S)}{a_3(X-X_S)+b_3(Y-Y_S)+c_3(Z-Z_S)} \\ y = -f\frac{a_2(X-X_S)+b_2(Y-Y_S)+c_2(Z-Z_S)}{a_3(X-X_S)+b_3(Y-Y_S)+c_3(Z-Z_S)} \end{cases} \quad (1)$$

Where *(X, Y, Z)* are the object coordinates, *(x, y)* are the coordinates in the image plane coordinate system, $(X_S, Y_S, Z_S)$ are the coordinates of the projection center. $a_i, b_i, c_i$ are the rotation matrix of three corner elements, and *f* is the focal length.

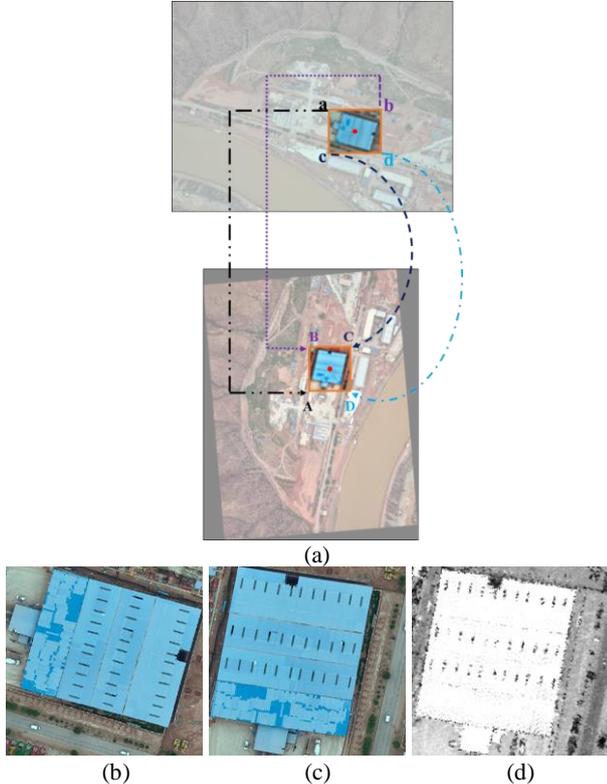

(a)

(b)　　　　　(c)　　　　　(d)

Figure 4. Processing chain of local geometric correction. (a) Schematic of local geometric correction. (b) Aerial image patch before local correction. (c) Aerial image patch after local correction. (d) LiDAR intensity image patch.

In the process of local geometric correction, we first select an image patch (i.e., a template region) centred on an interest point, and then employ the collinearity equation to correct the image patch. After t, the coarse registration is completed between two local image patches from the aerial and LiDAR intensity images. Figure 4 shows the processing chain of local geometric correction, we can see that the local correction effectively eliminates rotation and scale differences between the LiDAR and aerial data.

**2.3 Image matching using the similarity metric based on structural features and 3D phase correlation**

**2.3.1 Structural feature descriptor**: After the local geometric correction, the main problem for matching the LiDAR intensity and aerial images is to address the nonlinear intensity differences. Although the two types of images have quite different intensity information, similar structural properties can be clearly observed between them (Ye et al., 2017). Accordingly, structural features are used for the matching of the two types of images. Here, we employ a robust feature descriptor (named CFOG) to capture structural properties of images. CFOG is inspired from HOG, which is built by using orientated gradient information of images. Figure 5 illustrates the construction process of CFOG, which mainly include orientated gradient channels and 3D Gaussian convolution. The details of building CFOG are as follow.

Firstly, we calculate *m* orientated gradient channels for a given image, and uses $g_i$ to denote each orientated gradient channel, *1≤i≤m*. The orientated gradient channel is defined as:

$$g_o = \lfloor \partial I / \partial o \rfloor \quad (2)$$

where *I* is the given image, *o* is the gradient orientation, and ⌊ ⌋ means that it is equal to itself when its value is positive, otherwise zero.

In the practical calculation, it is not necessary to calculate the orientated gradient channel $g_o$ for each layer separately. Instead, the horizontal gradient $g_x$ and vertical gradient $g_y$ are used to calculate the orientated gradient channel for each layer. The calculation equation is as follows:

$$g_\theta = \lfloor abs(\cos\theta * g_x + \sin\theta * g_y) \rfloor \quad (3)$$

where *θ* is the orientation of orientated gradient, abs denote the absolute value which can limit the gradient orientation is in the range of [0°, 180°). This can address the intensity reverse between images.

Once completing the construction of the orientated gradient channel, we use a 3D Gaussian convolution kernel to achieve a convoluted feature channel by the following equation:

$$g_o^\sigma = g_\sigma * \lfloor \partial I / \partial o \rfloor \quad (4)$$

where σ is the standard deviation of the Gaussian convolution kernel. Strictly speaking, the kernel is not really 3D Gaussian function in the 3D space, but is a 2D Gaussian kernel in the X and Y directions and a kernel in gradient direction

$(1,2,1)^T$ (hereinafter referred to as the Z direction). The convolution in the Z direction can smooth the gradients in this direction, and can reduce the effect of orientation distortion caused by geometric deformation and intensity differences between images.

**2.3.2 Similarity evaluation**: CFOG is a 3D pixel-wise descriptor with large data volume, and it is quite time consuming if using the traditional similarity metrics (e.g., NCC and MI) for CP detection in the space domain. Since the local geometric correction (descripted in Section 2.2) can eliminate the differences in rotation and scale between two image patches, there are only translation shifts between them. A template match scheme is used for CP detection, where the 3D phase correlation is employed as the similarity metric in the frequency domain. This can effectively improve the computational efficiency because the dot production operation in the frequency domain corresponds to the correlation operation in the space domain.

Given a LiDAR intensity image $g(x, y, z)$ and aerial image $f(x, y, z)$, their corresponding 3D CFOG descriptors are $g_{CFOG}(x, y, z)$ and $f_{CFOG}(x, y, z)$, respectively. The translation relationship between them satisfies the following Equation:

$$g_{CFOG}(x, y, z) = f_{CFOG}(x - x_0, y - y_0, z) \quad (5)$$

Where $x_0$ is the horizontal offset, $y_0$ is the vertical offset, z is the dimension of the CFOG descriptor.

Let $3DG(u, v, w)$ and $3DF(u, v, w)$ be the 3D Fast Fourier transform (FFT) of the 3D CFOG descriptor of the LiDAR intensity and aerial images, respectively. According to the translation property of the Fourier transform, the relationship between $3DG(u, v, w)$ and $3DF(u, v, w)$ can be expressed as the Equation (6). Accordingly, the normalized cross power spectrum (Equation (7)) of the two CFOG descriptors is used as the similarity metric for image matching. Then, we perform the 3D inverse Fast Fourier transform (IFFT) of Equation (7) to obtain a 3D Dirac impulse function $H(x, y, z)$, which is expressed as Equation (8).

$$G(u, v, w) = F(u, v, w) e^{-i(ux_0 + vy_0)} \vec{\gamma} \quad (6)$$

$$H(u, v, w) = \frac{3DG(u,v,w)3DF(u,v,w)^*}{|3DG(u,v,w)3DF(u,v,w)^*|} = e^{-i(ux_0+vy_0)}\vec{\gamma} \quad (7)$$

$$H(x, y, z) = \vartheta^{-1}\{H(u, v, w)\} = \delta(u - x_0, v - y_0)\vec{\gamma} \quad (8)$$

Where $\vec{\gamma}$ is a 3D unit vector, $3DF(u, v, w)^*$ denote the complex conjugate of $3DF(u, v, w)$, $\vartheta^{-1}$ denote the 3D IFFT.

Dirac pulse function $H(x, y, z)$ can be used to determine the offset between two images (Foroosh et al., 2002). As shown in Figure 6, the pulse function has a sharp peak at the offset, and the maximum peak corresponds to the relative translation between the aerial and LiDAR intensity images. By determining the position of the peak, we can obtain the horizontal and vertical offsets $(x_0, y_0)$.

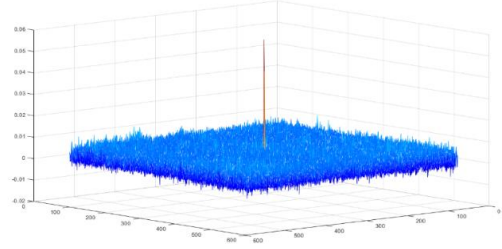

Fig 6. Pulse function peak

Figure 7 shows the matching process based on the structural features and 3D phase correlation, we employ a template match scheme for CP detection, which is as follows:

Step 1: Calculating the CFOG descriptor for a template region (i.e., image patch) of the aerial image using equations (3) and (4).

Step 2: The corresponding search region in the LiDAR intensity image is predicted according to the geographic information obtained by the local geometric correction, and the CFOG descriptor is also calculated for this region.

Step 3: Performing the 3D FFT for the obtained CFOG descriptors and calculating the normalized cross power spectrum between the two template regions using Equation (7).

Step 4: Performing the 3D IFFT to yield a Dirac impulse function using Equation (8), and finding its maximum to achieve the offset $(x_0, y_0)$ between the two template regions.

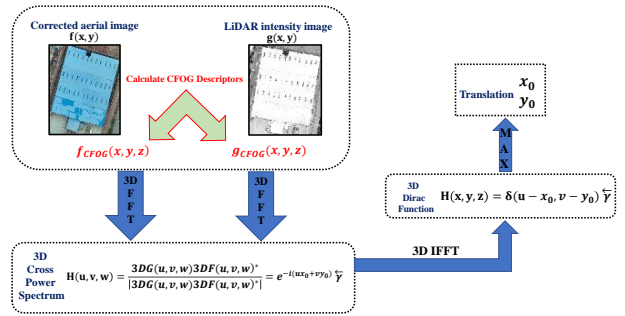

Figure 7. Matching process based on structure features and 3D phase correlation

In order to illustrate the matching advantage of the proposed similarity metric for aerial and LiDAR intensity images, the proposed similarity metric is compared with the two classic similarity metrics such as NCC and MI. As shown in Figure 8, a pair of aerial image and LiDAR intensity image is used as the test data for this comparison, in which the template window size is 80×80 pixels and the search window size is 60×60 pixels. The similarity of these similarity metrics is represented by different colors, where the blue indicates low similarity, and the red denotes high similarity.

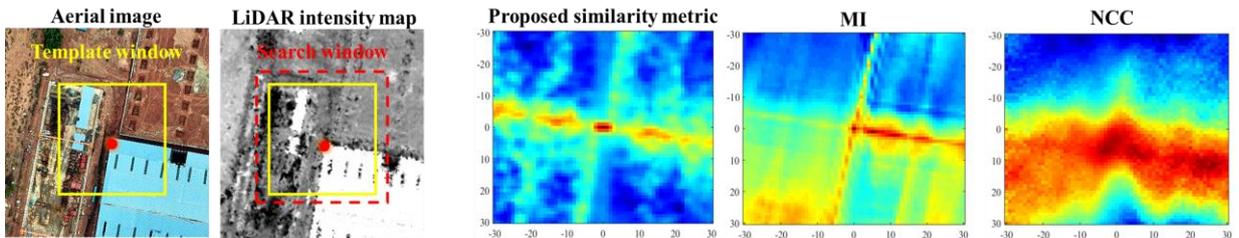

Figure 8. Similarity Map

| Test case | Dataset description | | |
|---|---|---|---|
| | LiDAR intensity image | Aerial image | Image characteristic |
| Test 1 | Resolution: 0.2m<br>Data: September, 2018<br>Size: 9128*9436 | Resolution: 0.2m<br>Data: September, 2018<br>Size: 10336*7788 | The images cover the suburban areas with buildings, farmlands, forests. There is the obvious noise on the LiDAR intensity image. |
| Test 2 | Resolution: 0.2m<br>Data: May, 2019<br>Size: 2837*3582 | Resolution: 0.03m<br>Data: May, 2019<br>Size: 11608*8708 | The images cover the suburban areas with buildings, rives and mountains. Moreover, there is significant intensity differences between them. |

Table 1. Detailed description of test cases

We can see from Figure 8 that NCC fails in the image matching because its peak dose not locate the correct position. Although MI has a sharper peak, the peak also has some location errors. The peak of the proposed similarity metric is smoother and corresponds to the correct matching position. This test preliminarily indicates that the proposed similarity metric is better than MI and NCC for the matching of aerial and LiDAR intensity images. The more experimental analysis of the proposed similarity metric will be given in Section 3.

**2.4 Exterior orientation element optimization and outlier detection**

By the approaches mentioned above, it can obtain a certain number of CPs between aerial and LiDAR intensity images. The geographical coordinates (X, Y, Z) of CPs can be obtain from the LiDAR intensity and elevation images. Then, these CPs are used to perform the resection based on the collinearity equation to achieve more accurate exterior orientation elements for aerial images.

Considering that the obtained CPs inevitably have some errors, we apply the following steps to remove these mismatches. (1) exterior orientation elements of aerial images are computed based on the CPs by using the least square method. (2) The residuals and root-mean-square errors (RMSEs) of these CPs are calculated, and the CPs with large errors are removed. (3) The above two steps are repeated until the RMSE is less than a certain threshold. Once mismatches are removed, we use the correct CPs to optimize exterior orientation elements of aerial images, and perform image registration for aerial and LiDAR intensity images.

## 3. EXPERIMENTS

To verify the performance of the proposed method, it is compared with the two state of art methods (i.e., NCC and MI). Same as the proposed method, NCC and MI are also used for image registration with the local geometric correction (descripted in Section 2.2) to make a fair comparison. These methods are evaluated using three criteria, which are the correct match number (CMN), the RMSE and the running time. In the experiments, the LiDAR intensity image is used as the reference image and the aerial image is used as the input image. The parameters of image matching are set as follows. The aerial image is divided into 20×20 grids, and one interest point is extracted in each grid, reaching a total of 400 interest points. The template size is set to 200×200 pixels and the search size is set to 100×100 pixels. The experiments are performed using a PC with a configuration of Inter (R) Core (TM) CPU i5-5200U 2.2GHz and 4GB RAM.

**3.1 Datasets**

The two pairs of aerial and LiDAR intensity images are selected as the experimental data. The details of experimental test are shown in Table 1. These images covering the suburban areas, which includes buildings, farmlands, forests, and roads (see Figure 9). It can be clearly observed that significant geometric deformation and intensity differences exists between the two images, which make it quite challenging to match them.

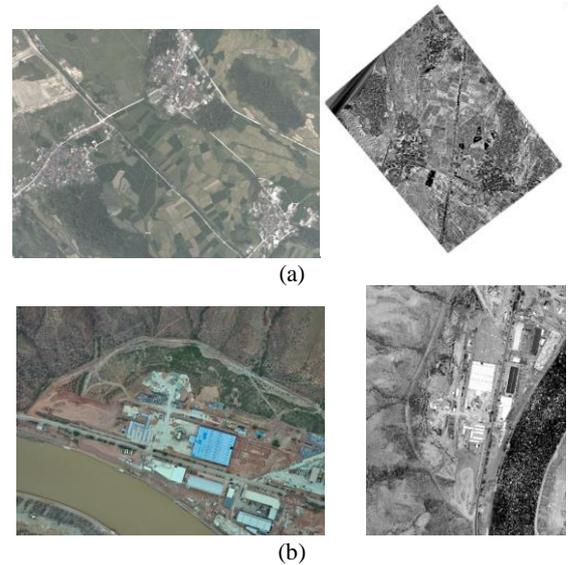

Figure 9. Experimental data. (a) Aerial and LiDAR intensity images of test 1. (b) Aerial and LiDAR intensity images of test 2.

| Test | Method | CMN | RMSE (pixels) | Running Time (second) |
|---|---|---|---|---|
| Test 1 | Proposed method | 138 | 1.876 | 120.460 |
| | MI + local correction | 79 | 2.212 | 5781.227 |
| | NCC + local correction | 41 | 2.421 | 1838.392 |
| Test 2 | Proposed method | 113 | 1.903 | 115.583 |
| | MI + local correction | 71 | 2.356 | 5699.167 |
| | NCC + local correction | 37 | 2.497 | 1725.248 |

Table 2. Registration results of the proposed method, MI, and NCC

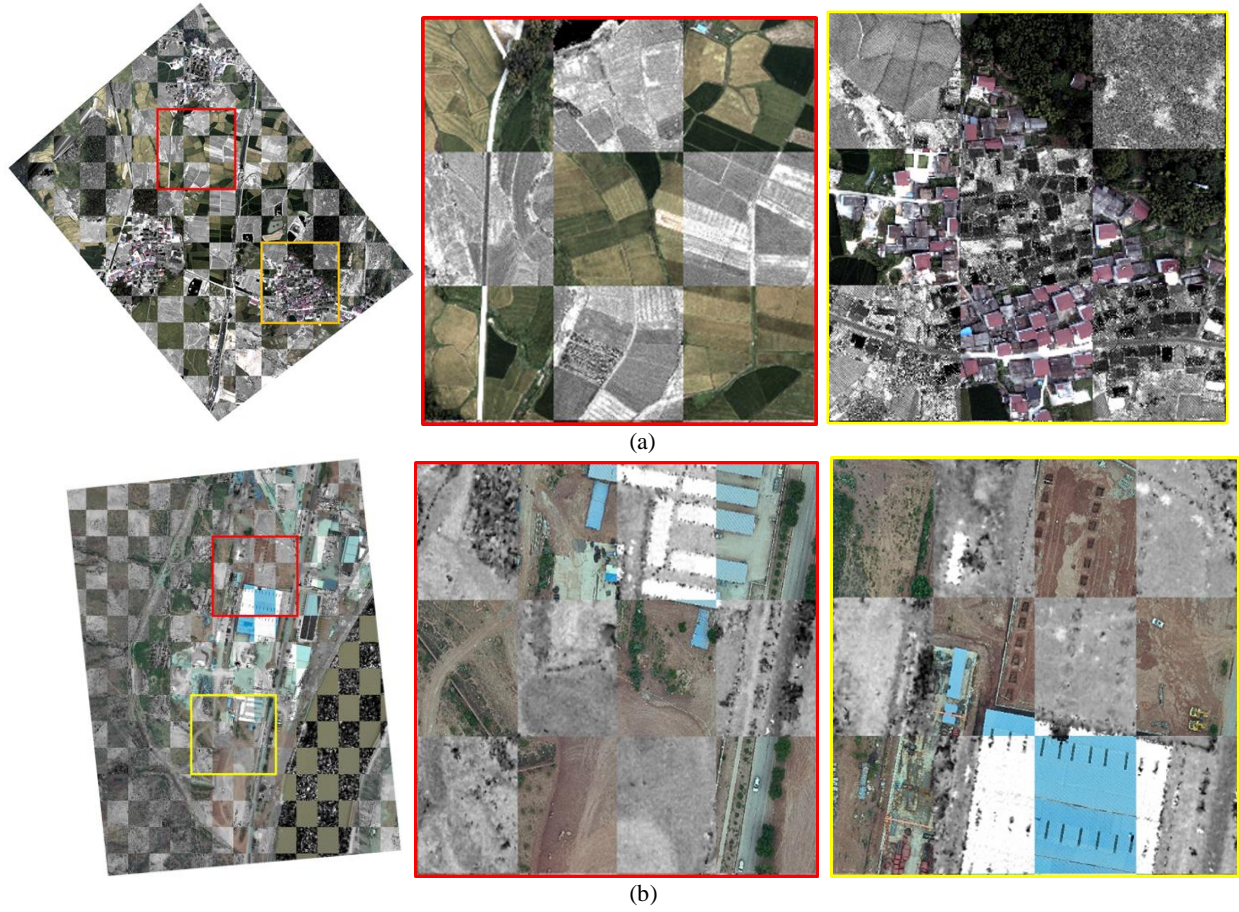

Figure 10. Checkboard visualization of registration results. (a) Checkboard map of the registration result of test 1. (b) Checkboard map of the registration result of test 2.

| Test | elements | Initial | Correction | Final |
|---|---|---|---|---|
| Test 1 | $X_s$/m | 484249.97 | -11.75 | 484261.72 |
| | $Y_s$/m | 3328438.95 | -5.13 | 3328433.82 |
| | $Z_s$/m | 4017.69 | -34.76 | 3982.93 |
| | Phi/° | -0.22 | -0.85 | -1.07 |
| | Omega/° | -0.26 | 0.53 | 0.27 |
| | Kappa/° | 94.72 | 1.35 | 96.07 |
| Test 2 | $X_s$/m | 545061.49 | -27.88 | 545033.61 |
| | $Y_s$/m | 3266654.56 | -4.39 | 3266650.17 |
| | $Z_s$/m | 1476.03 | -4.91 | 1471.12 |
| | Phi/° | -0.97 | 3.06 | 2.09 |
| | Omega/° | 4.16 | 0.17 | 4.33 |
| | Kappa/° | 129.25 | -0.1 | 129.15 |

Table 3. Optimization of exterior orientation elements

### 3.2 Analysis of accuracy and computational efficiency

Table 2 shows the registration results of the three methods in terms of CMN, RMSE and running time. We can see that NCC achieves the least CMNs and the lowest registration accuracy. It is attributable to that NCC can only handle linear intensity changes, and is vulnerable to nonlinear intensity differences between images. Compared with NCC, the CMNs and RMSEs of MI have some improvements because MI can resist nonlinear intensity differences to some extent. However, MI yields many mismatches and its registration accuracy is not quite satisfactory. In comparison to NCC and MI, the proposed method achieves more CMNs and higher registration accuracy. This is because the proposed method employs structural features for image registration, which is robust to significant nonlinear intensity differences.

In the running time, MI is the most time consuming, because it needs to calculate a large number of joint probability histograms between templates (Parmehr et al., 2014). NCC takes the less running time than MI, but it achieves the least CMNs and the lowest registration accuracy. In contrast, the proposed method is about 50 times and 15 times faster than MI and NCC, respectively. This is because the propose method performs image matching using the 3D phase correlation in the frequency domain, which can effectively improve the computational efficiency. In a word, the proposed method outperforms NCC and MI in both registration accuracy and computational efficiency. Figure 10 shows the registration results. We can see that the aerial image and the LiDAR intensity image have been aligned exactly.

Due to the differences between the two datasets, test 1 and test 2 present different registration accuracy. The registration accuracy of test 1 is slightly higher than that of test 2, which may be attribute to the fact that the resolution of the images of test 2 is higher than that of test 1. It's means that the images of test 2 have more significant local geometric distortions, which results in the degradation of the registration performance.

Table 3 shows the results of optimizing the exterior orientation elements for the two tests. The proposed method makes use of the georeference information in LiDAR data to achieve more accurate exterior orientation elements for the aerial images.

## 4. CONCLUSIONS

This paper presents an automatic registration method based on structural features and 3D phase correlation, in order to address significant geometric distortions and nonlinear intensity differences between the aerial and LiDAR intensity images. The proposed method first performs a local geometric correction to eliminate the rotation and scale differences between images. Then 3D phase correlation based on structural features are used as the similarity metric to achieve CPs between images. Finally, These CPs are used to optimize the exterior orientation elements of aerial images, and perform image registration. Two pairs of aerial and LiDAR data have been used to evaluate the proposed method. Experimental results show that the proposed method outperform the state of the art methods (i.e., NCC and MI), and can achieve reliable registration accuracy for aerial images and LiDAR data.

The main limitation of the proposed method is that it relies on rich structure properties of images. The poor structure information will degrade the registration performance. In addition, the local geometric correction based on the collinearity equation is the precondition of the proposed method. These limitations will be addressed in future works.

## ACKNOWLEDGEMENTS

This paper is supported by the National Natural Science Foundation of China (No.41971281).